%% file: main.tex
\documentclass[runningheads]{llncs}

\usepackage{eccv}

\usepackage{eccvabbrv}

\usepackage[accsupp]{axessibility}  % Improves PDF readability for those with disabilities.

\usepackage{hyperref}

\usepackage{orcidlink}

\usepackage{graphicx}
\usepackage{booktabs}
\usepackage{makecell}
\usepackage{multirow}
\usepackage{adjustbox}
\usepackage{threeparttable}

\usepackage{tikz}
\usetikzlibrary{positioning, fit}

\pgfdeclarelayer{background}
\pgfsetlayers{background,main}

\usepackage{comment}
\usepackage{caption} 
\usepackage{subcaption}

\begin{document}

%\title{Finding the eye: an improved ground truth for eye tracking dataset} 
\title{High-frequency near-eye ground truth for event-based eye tracking}

\titlerunning{High-frequency near-eye ground truth for event-based eye tracking}

\author{Andrea Simpsi\inst{1}\thanks{Corresponding author.}\orcidlink{0000-0002-2700-4808} \and
Andrea Aspesi\inst{1,2}\orcidlink{0009-0000-4936-5973} \and
Simone Mentasti\inst{1}\orcidlink{0000-0001-7059-413X} \and
Luca Merigo\inst{2}\orcidlink{0000-0001-9103-1774} \and
Tommaso Ongarello\inst{2} \and
Matteo Matteucci\inst{1}\orcidlink{0000-0002-8306-6739}}

\authorrunning{A.~Simpsi et al.}

\institute{
    Department of Electronics, Information and Bioengineering (DEIB)\\
    Politecnico di Milano\\
    Via Ponzio 34/5, 20133 Milan, Italy\\
    \email{\{name.surname\}@polimi.it}
    \and
    EssilorLuxottica Italia S.p.A.\\
    Piazzale Cadorna 3, 20123 Milan, Italy\\
    \email{\{name.surname\}@luxottica.com}
}

\maketitle

\begin{abstract}
 Event-based eye tracking is a promising solution for efficient and low-power eye tracking in smart eyewear technologies. However, the novelty of event-based sensors has resulted in a limited number of available datasets, particularly those with eye-level annotations, crucial for algorithm validation and deep-learning training. This paper addresses this gap by presenting an improved version of a popular event-based eye-tracking dataset. We introduce a semi-automatic annotation pipeline specifically designed for event-based data annotation. Additionally, we provide the scientific community with the computed annotations for pupil detection at $200Hz$.
\end{abstract}

\section{Introduction}
\input{content/1-introduction}

\section{Related Works}
\input{content/2-soa}

\section{Ground truth semi-automatic generation}
\input{content/3-work}

\section{Dataset statistics}
\input{content/4-metrics}

\section{Conclusions}
\input{content/5-conclusions}

\section*{Acknowledgments}
This paper was carried out in the EssilorLuxottica Smart Eyewear Lab, a Joint Research Center between EssilorLuxottica and Politecnico di Milano.

%\clearpage\mbox{}Page \thepage\ of the manuscript.
%\clearpage\mbox{}Page \thepage\ of the manuscript.
%\clearpage\mbox{}Page \thepage\ of the manuscript.
%\clearpage\mbox{}Page \thepage\ of the manuscript.
%\clearpage\mbox{}Page \thepage\ of the manuscript. This is the last page.
%\par\vfill\par
%Now we have reached the maximum length of an ECCV \ECCVyear{} submission (excluding references).
%References should start immediately after the main text, but can continue past p.\ 14 if needed.
\clearpage  % TODO REVIEW/FINAL: This \clearpage needs to be removed from both review and camera-ready versions.

\bibliographystyle{splncs04}
\bibliography{main}
\end{document}

%% file: content/1-introduction.tex
Eye tracking is a technology that measures and records the eye movements and positions of an individual~\cite{andrychowicz2018basic}, providing valuable insights into visual attention and cognitive processes~\cite{Popa2015Reading}~\cite{Skaramagkas2021Review}. This technology has profound implications for various applications, ranging from user experience research~\cite{Burger2018Suitability} and medical diagnostics~\cite{Liu2021The} to advanced human-computer interaction systems~\cite{Bulling2010Toward}~\cite{Zhang2017Eye}. Moreover, with the advent and rapid diffusion of virtual reality devices~\cite{angelov2020modern}, augmented reality systems~\cite{arena2022overview}, and smart eyewear, eye-tracking technology has become a central component in these systems to enhance the user experience and provide a more immersive and interactive environment.

Thanks to the proliferation of smart wearables and smart eyewear in recent years~\cite{spil2019adoption}, research on this topic has witnessed a significant boost. In particular, in recent years, we have seen rapid growth in the availability of commercial eye-tracking technologies~\cite{KassnerPateraBulling}. These devices, equipped with advanced sensors, enable continuous and unobtrusive monitoring of eye movements in real-world settings. This has opened up new possibilities for applications such as biometric authentication, emotion recognition, and user experience optimization.

Despite the recent advancements, these new technologies still struggle to provide real-time information, or they require heavy computational power and high-end PCs to process the data~\cite{wan2021robust}. Recent advancements in computer vision have significantly enhanced the accuracy of eye-tracking technology. In particular, machine learning and deep learning approaches have enabled more accurate and robust detection and tracking of eye movements, even in challenging conditions such as varying lighting and head movements~\cite{zhu2007novel}. Techniques such as convolutional neural networks (CNNs) and recurrent neural networks (RNNs) have been instrumental in improving the precision and speed of eye tracking systems~\cite{fuhl2016pupilnet}~\cite{santini2018pure}. Nevertheless, the improved accuracy has come at the cost of using powerful GPUs to run the deep-learning models, making the whole eye-tracking system more resource-demanding and distant from a slim eye-wear system, where all the computation is performed directly on the device.

The growing accuracy in the eye tracking process and the increase in computational requirements are related to how traditional eye tracking has been performed, using cameras looking at the user's eye. Traditionally, an eye-tracking pipeline is divided into a series of refinement blocks that gradually find the pupil in the image, then the center, and track it through time. It is then expected that an increase in accuracy can lead to higher computational requirements. For this reason, in recent years, alternative solutions in eye tracking have been proposed, from non-camera based like microelectromechanical systems (MEMSs) to scan the eyes~\cite{zafar2023investigation} to photodetector-based solutions~\cite{crafa2024towards}. Other solutions focused on new and innovative camera systems, particularly event-based cameras~\cite{gallego2020event}, which, unlike traditional frame-based sensors, return only the changing pixels from one frame to the next, in an asynchronous way. This feature can be exploited to detect only the pupil in the frame and drastically increase the performance of the pupil detector while maintaining low computational power.

In particular, event-based eye tracking has proved to be a promising solution. However, annotated data with proper ground truth is required to validate eye-tracking algorithms and train deep-learning models. Camera-based eye-tracking datasets are fairly common and have been recorded for many years~\cite{winkler2013overview}. In contrast, to the best of our knowledge, the number of datasets for these new sensors is extremely limited due to the novelty of the event-based approach. Indeed, only four datasets are available~\cite{angelopoulos2020event}~\cite{bonazzi2024retina}~\cite{wang2024eventbasedeyetrackingais}~\cite{eveyepaper}, but each of them presents some limitations that are further discussed in the following sections.

\begin{comment}Indeed, only two datasets are available~\cite{angelopoulos2020event}~\cite{bonazzi2024retina}. A third one exists ~\cite{wang2024eventbasedeyetrackingais}, yet was collected without IR-pass filter so it contains events of object reflections mixed with eye movements.

In Angelopoulos, the dataset has limited ground truth, due to the difficulties in generating a pupil-level annotation from the sparse event data. Indeed, Angelopoulos's dataset provides as ground truth the looked point on a screen, not the pupil position in the image frame. This type of annotation has been widely used for many years in the eye-tracking field~\cite{holmqvist_eye_movements} but introduces additional calibration and nonlinearities in the system, which makes a proper evaluation of the pure pupil detection and tracking algorithm less accurate.
\end{comment}

For this reason, in this work, we present an improved version of the original dataset proposed by Angelopoulos in~\cite{angelopoulos2020event}. In particular, we present a pipeline for semi-automatic annotation of event data for pupil detection and tracking. We also provide the scientific community with the generated pupil position for Angelopoulos's dataset at 200Hz. The computed annotations are made publicly available at \footnote{\url{https://github.com/AIRLab-POLIMI/event_based_gaze_tracking_gt}}.

This work is organized as follows: Section~\ref{sec::soa} presents the current state of the art in eye-tracking, with a specific focus on the event-based scenario and the datasets available. Subsequently, Section~\ref{sec::ann} describes our proposed pipeline for automatic annotation generation from event data. In Section~\ref{sec::eval}, we highlight key statistics regarding the size of the annotations generated. Finally, Section~\ref{sec::end} offers concluding remarks and summarizes the findings of this paper.

%% file: content/2-soa.tex
\label{sec::soa}
In this section, we discuss the datasets that provide event-camera data specifically for eye-tracking applications. As previously anticipated, due to the novelty of the sensor and application, the number of available datasets is extremely low.

%\subsection{Angelopoulos} \label{angelopoulos_soa}
%todo~\cite{angelopoulos2020event}
One of the first datasets for the eye-tracking task is the one presented by Angelopoulos in~\cite{angelopoulos2020event}. This dataset has been created using data obtained by the DAVIS sensor, which provides both greyscale and event-based data. The dataset is composed of several eye movements obtained from 27 different users. 
The ground truth provided is unfortunately limited due to the difficulties in generating a pupil-level annotation from the sparse event data. Indeed, Angelopoulos's dataset provides as ground truth the looked point on a screen, not the pupil position in the image frame. This type of annotation has been widely used for many years in the eye-tracking field~\cite{holmqvist_eye_movements} but introduces additional calibration and nonlinearities in the system, which makes a proper evaluation of the pure pupil detection and tracking algorithm less accurate.

%\subsection{Retina} \label{retina_soa}
%todo~\cite{bonazzi2024retina}
A recent addition is the Ini-30 dataset~\cite{bonazzi2024retina}, obtained by recording 30 volunteers using a glass frame equipped with two DVXplorer event cameras. To the best of our knowledge, this is the first event-based dataset where labels are provided at high-frequency directly at frame level and where a screen was not employed to guide the users during the acquisition process and with labels dire. The glasses cameras captured natural eye movements; then, data were manually labeled with a variable sampling period ranging from 20.0ms to 235.77ms. The resulting 50Hz labeling frequency, however, is not high enough to capture all eye movements, as presented in the next sections. Additionally, the absence of a standardized protocol during acquisitions, although enhancing variability, might lead to imbalances in the data, thereby affecting its usability.

%\subsection{EV-Eye} \label{eveye_soa}
%todo~\cite{eveyepaper}
Another dataset was recently presented as part of the EV-Eye paper \cite{eveyepaper}. Similarly to \cite{angelopoulos2020event}, two DAVIS346 are used to capture events and also record near-eye grayscale image sequences at a frame rate of 25fps. In addition, it contains gaze references provided by Tobii Pro Glasses 3. This additional metadata comprises PoGs and pupil diameters of the users at $100Hz$. The dataset is composed by 48 participants (28 male and 20 female) aged between 21 and 35 years. Labels are also provided at sensor level as in~\cite{bonazzi2024retina}, but only at slow frequency as estimated from near-eye grayscale images.

%\subsection{AIS 2024 Challenge} \label{challenge_soa}
%todo \cite{wang2024eventbasedeyetrackingais}
Recently a new event-based dataset has also been presented, the 3ET+ dataset~\cite{wang2024eventbasedeyetrackingais}. In this scenario the data have been obtained using a DVXplorer Mini event camera. The recording consists of 13 distinct users, each exhibiting various eye movements. Unlike the previous datasets, 3ET+ provides the ground truth annotated at 100Hz. Moreover, they provide two different labels: one binary value to indicate the blink status, and the human-labeled pupil center coordinates. However, the data were collected without an IR-pass filter, so they contain events of object reflections mixed with eye movements. This could impact the generalization capability of deep learning algorithms trained on this dataset.

This work advances the current state of the art by providing additional annotations to the Angelopoulos dataset~\cite{angelopoulos2020event}. Specifically, we retrieve the center of the pupil on the image frame at a rate of 200 Hz, a frequency sufficiently high to model all documented eye movements~\cite{holmqvist_eye_movements}. Additionally, we include annotations describing the current state of the eye, labeling the presence of a blink and the status of the saccade.

%% file: content/3-work.tex
\label{sec::ann}
In this section, we describe the methodology used to generate pupil position and blink annotations from Angelopoulos's dataset~\cite{angelopoulos2020event}. First, the annotations are computed using an automated process. These annotations are subsequently validated and, if necessary, corrected through manual verification.

\begin{figure}[t]
  \centering
  \includegraphics[width=0.9\textwidth]{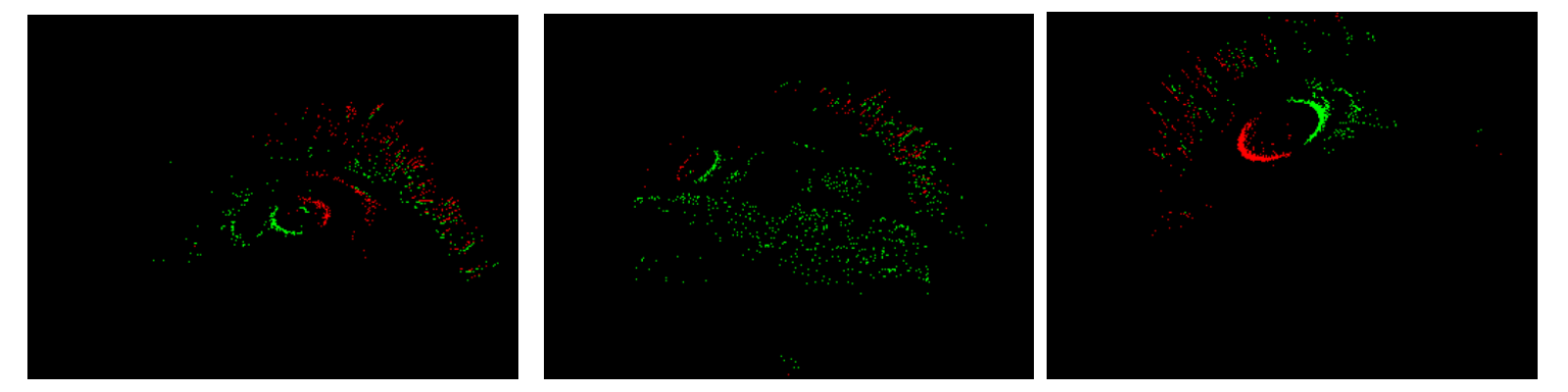}
  \caption{Examples of frames generated accumulating events at 200Hz}
  \label{fig:Polarity_frame}
\end{figure}

\subsection{Automatic annotation } \label{automatic_annotation }
A notable drawback of event cameras is the difficulty of directly applying traditional frame-based algorithms to their output data. This issue arises because event cameras generate data as a continuous stream of asynchronous events. Unlike frame-based cameras, event cameras only capture changes in individual pixels, without producing complete images. This limitation impacts the data visualization and the annotation process. Therefore, the first step in the automatic annotation pipeline is to generate RGB frames from the event stream. These frames possess all the characteristics of traditional camera images, allowing them to be visualized and annotated by both the automatic pipeline and the users.

\begin{figure}
\centering
\includegraphics[width=0.80\textwidth, trim=80px 19px 170px 0px, clip]{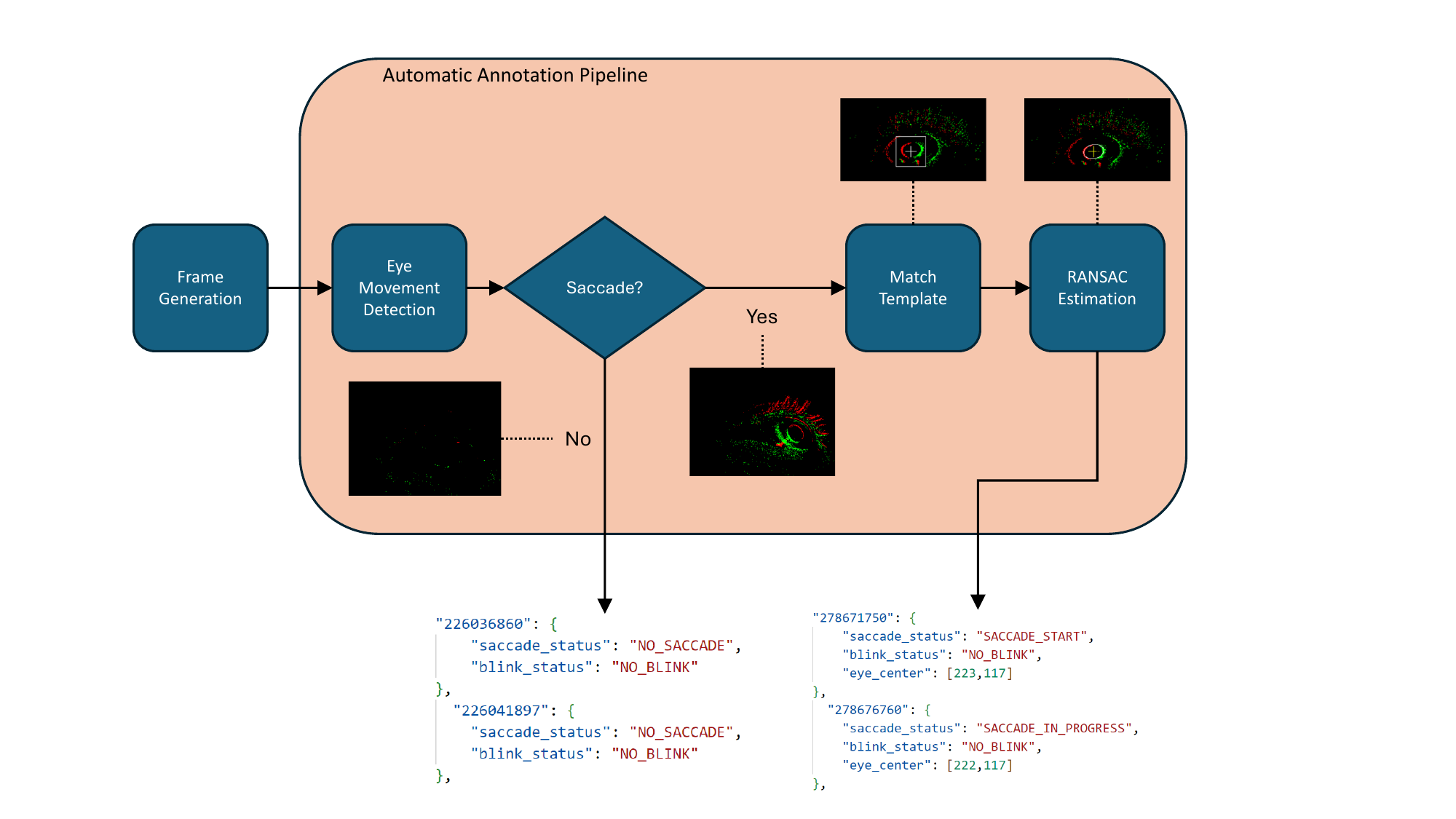}
\caption{Schema of the automatic annotation pipeline. The system takes the frame generated by accumulating events as input and first predicts if there has been eye movement. If a saccade is detected, the pupil center is determined using a template matching strategy followed by RANSAC estimation.}
\label{fig:auto_annotation_pipeline}
\end{figure}

To generate the event frames, we aggregated events at a sampling rate of $200 Hz$ (i.e., a frame every $5ms$). This means that all events, which represent changing pixels, in a $5ms$ window are summed to reconstruct an RGB image of the eye.  This sampling rate was selected to capture all relevant eye movements. Indeed, saccadic movements typically span from $30 ms$ to $80 ms$, while microsaccadic movements occur within a $10 ms$ to $30 ms$ range~\cite{holmqvist_eye_movements}. The need for such a high sampling frequency to accurately capture all types of eye movements introduces substantial challenges in the annotation process. As the sampling frequency increases, there is a proportional increase in the volume of frames that require annotation, complicating the data processing workflow.

The result of the accumulation process are RGB frames where the only colored points are the events accumulated during the sampling period. The color of the points indicates the polarity of each event: green for positives and red for the negative ones. The output of this first phase of the pipeline is shown in Fig.~\ref{fig:Polarity_frame}.

The next phase involves automatically generating the raw annotations for each frame. Due to the high frequency at which the data are processed, a significant portion of the frames contain limited information, with most colored pixels attributable to noise. This phenomenon occurs when the eye is nearly stationary; if no movement is detected by the sensor, no events are generated. In these scenarios, where only noise data are available, automatic annotation is problematic due to the lack of a clear pupil. In this scenario, manual inspection would only increase the effort required by the human operator without adding additional benefit to the final result.

%Then, we start by generating a base annotation using an automatic process, in order to reduce manual rework to a minimum. Most of the frames, at this frequency, contain noise and not eye movements, so it would not be feasible to manually inspect these frames. 

%The automatic annotation pipeline is shown in Fig.~\ref{fig:auto_annotation_pipeline}, and it's divided into three major steps:
%\begin{enumerate}
%    \item We need to detect if either a saccade or blink is ongoing.
%    \item If a saccade is ongoing, we find a tentative eye center using a pattern-matching methodology.
%    \item We further improve the center using RANSAC fit on the events contained in a ROI centered on the tentative center.
%\end{enumerate}
%These three steps utilize the method described in~\cite{mentasti2024event}. The saccade classification is performed using a running mean with the same parameters outlined in the article. Similarly, the pattern-matching methodology employs a slightly improved kernel compared with the one presented in that study. Although the RANSAC fit remains the same, it uses a higher number of iterations in this scenario: 1000 iterations.

\begin{figure}[t]
    \centering
    
    \begin{subfigure}[t]{0.32\textwidth}
        \centering
        \includegraphics[width=\linewidth]{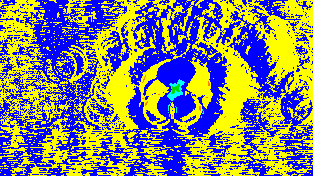}
        \caption{Result of a convolution from match template phase, the light blue zone indicates where the pupil can be located.}
        \label{fig:convolutions}
    \end{subfigure}
    \hfill % optional, for better spacing
    \begin{subfigure}[t]{0.32\textwidth}
        \centering
        \includegraphics[width=\linewidth]{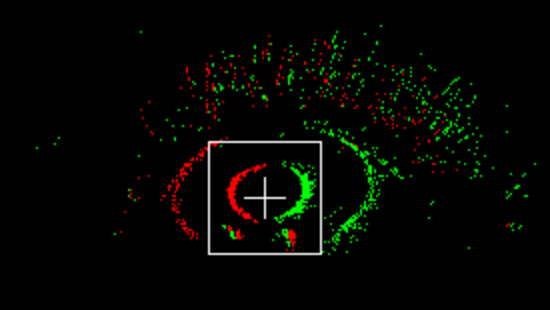}
        \caption{Expected pupil center after the match template step.}
        \label{fig:tentative_center}
    \end{subfigure}
    \hfill % optional, for better spacing
    \begin{subfigure}[t]{0.32\textwidth}
        \centering
        \includegraphics[width=\linewidth]{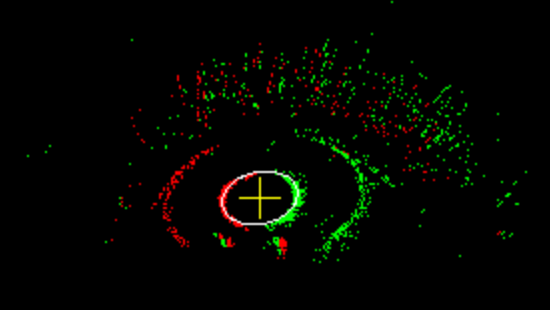}
        \caption{Estimated pupil center after the RANSAC estimation step.}
        \label{fig:ransac_fit}
    \end{subfigure}
    
    \caption{Results from different operations performed during the pupil center localization. As we can see, the RANSAC estimation step increases the accuracy of the estimation made by the match template step.}
    \label{fig:predictor_3g}
\end{figure}

The automatic annotation pipeline is shown in Fig.~\ref{fig:auto_annotation_pipeline}, and it is divided into three major steps inspired by the method described in:~\cite{mentasti2024event}:
\begin{itemize}
\item \textbf{Eye Movement Detection}: In this step, we need to detect if either a saccade or a blink is ongoing. We can detect the presence of a saccade using a threshold technique. If the events in the frame are more than 150, there is a saccade; otherwise, it is not a saccade. Also, the first time that a frame is detected as a saccade, it is labeled as \textit{SACCADE\_START}. Similarly, the frame is labeled as \textit{SACCADE\_END} if it is the last frame before switching to the frames without detected saccades.
\item \textbf{Match Template:} If a saccade is ongoing, the match template step finds a tentative eye center using a pattern-matching methodology. To do this, we use a 2D convolution similar to~\cite{mentasti2024event}, but with a slightly improved kernel compared with the one presented in the original work. In particular, to improve system performance, we employ only 8 different templates that represent different movement directions. These kernels are applied to the polarity images (i.e., the frames reconstructed integrating events data). The output of this phase is shown in Fig.~\ref{fig:convolutions}. Then, each result is multiplied to create heatmaps, identifying movement direction. The highest-scoring template indicates the motion direction; a bounding box, representing the Region Of Interest (ROI) of the frame, is built centered at the maximum point of the heatmap. The result is visible in~\ref{fig:tentative_center}.
\item \textbf{RANSAC Estimation}: After the match template step, we execute the Random Sample Consensus (RANSAC) algorithm~\cite{fischler1981random}, considering only the events inside the ROI detected earlier. In particular, we employ RANSAC to fit an ellipse described using the equation of a generic conic with the f parameter fixed to -1. Although the RANSAC fit follows the method described in~\cite{mentasti2024event}, it uses 1000 iterations in this scenario to prioritize reliable results over fast execution. The final result is visible in~\ref{fig:ransac_fit}.
\end{itemize}

\subsection{Fine Tuning Annotation using Human Annotators} \label{fine_tune_annotation}

After generating and automatically annotating each frame, as described in the previous section, the labels are inspected by human annotators with expertise in eye-tracking and event cameras to validate and correct the annotations.

The annotation fine-tuning process is divided into two parts. The first part aims to assign each frame a specific label describing the state of the eye, such as a saccade or blink. In this stage, each annotator reviews the automatic annotations generated in the previous step. Annotators are presented with each frame that contains a number of events exceeding a pre-set threshold, which varies between users. This method excludes all frames with a low number of events, where the eye's state is not clearly defined, and the human annotator would be unable to provide accurate additional information.

Using a simple keyboard shortcut-based interface, each annotator is able to modify the eye center's position, to change the saccade status (\textit{SACCADE\_START}, \textit{SACCADE\_IN\_PROGRESS}, \textit{SACCADE\_END}) and select the current blink status (\textit{BLINK\_STATUS}, \textit{BLINK\_IN\_PROGRESS}, \textit{BLINK\_END}). Having a base annotation greatly speed-up the review process, as for many frames the base annotation itself is correct.

\begin{figure}[t]
  \centering
  \includegraphics[width=0.9\textwidth]{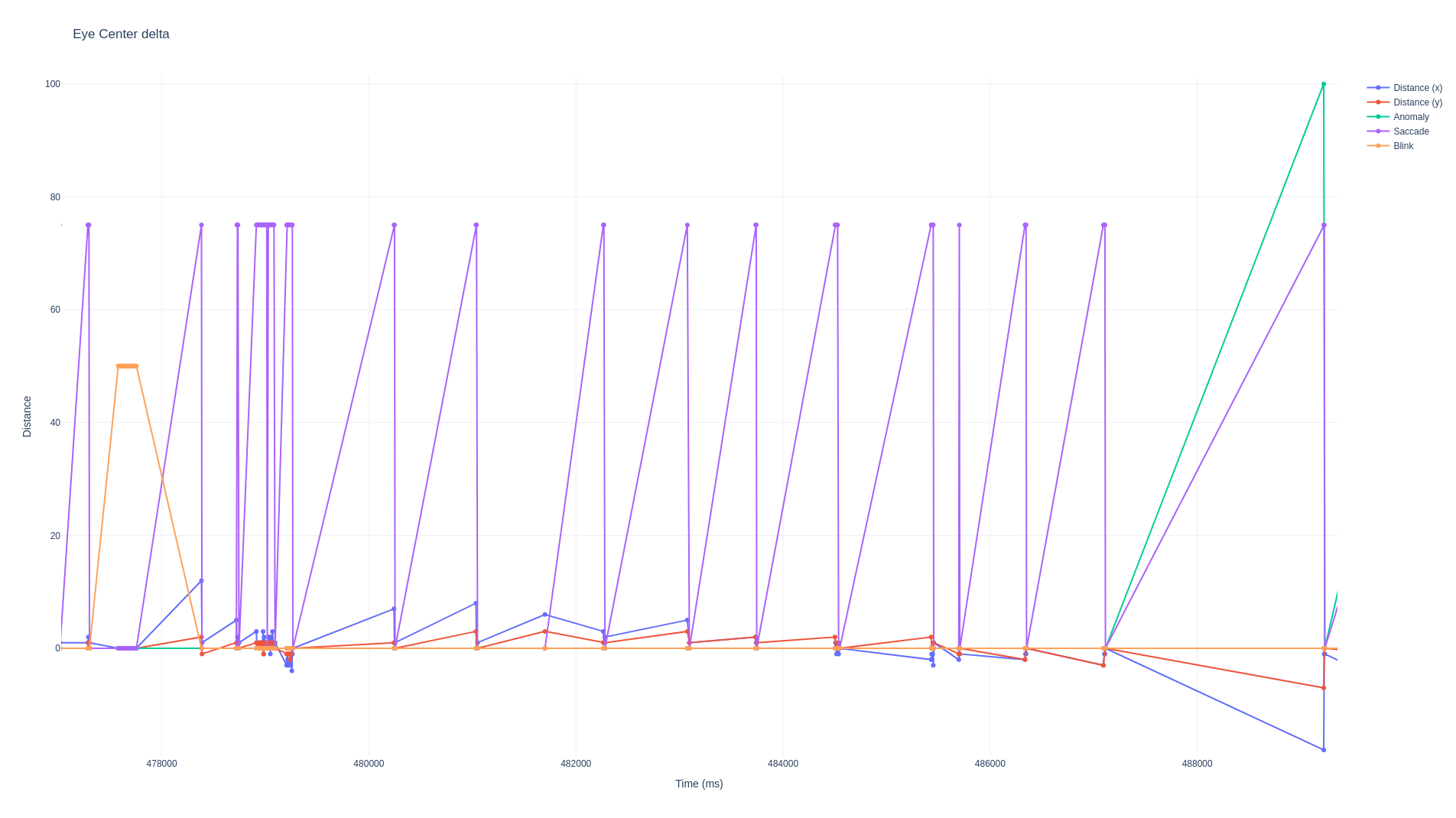}
  \caption{Interactive plot for annotation correction. Delta in x,y from an eye position to the next is shown in blue and red. Saccades are marked in violet (rising edge when the saccade starts, falling edge when the saccade ends). Blinks are indicated similarly in orange. At the end of the sequence, a possible anomaly is marked in green.}
  \label{fig:anomaly_detection}
\end{figure}

In the second stage, the quality of the labels is evaluated using another custom-made tool. Considering the high frequency of annotations, we expect extremely small eye movements to occur between frames. To leverage this, we introduced a metric based on the difference in eye position between consecutive frames. A threshold for this distance is set to identify potential label anomalies, under the assumption that pupil movement should not exceed this distance within a $5 ms$ interval. The output from this phase is displayed as an interactive plot (see Fig.~\ref{fig:anomaly_detection}). Reviewers can then zoom into specific plot regions where the threshold is breached and directly jump to the corresponding labels for correction.

This second step is performed iteratively until the plot no longer exhibits any significant anomalies. Some peak distances may persist due to eye movements occurring during blinks. The presence of these peaks is largely attributed to different physiological mechanisms; for example, downward eye movements during blinks are well-documented in the literature~\cite{Khazali2017}. Moreover, saccades and blinks can be simultaneous.

%% file: content/4-metrics.tex
\label{sec::eval}
Using the method proposed in this paper, we annotated the left eye of six different users from the dataset~\cite{angelopoulos2020event}.

We reviewed and annotated 114222 frames, each containing a varying number of events. The total number of frames reviewed was 315213; however, many did not contain the minimum number of events necessary to recognize and detect an eye. These empty frames likely resulted from periods when the eyes were still and did not generate events.

\begin{table}[t]
\centering
    
\resizebox{\textwidth}{!}{
\setlength{\tabcolsep}{10pt}
\begin{tabular}{c|ccccccc|c|}
 \hline
 \hline
 Statistic& User4 & User5 & User10 & User15 & User18 & User19 & User20 & Total\\
 \hline
 Frame analyzed   & 44020 & 41233 & 46818 & 46008 & 43470 & 43125 & 50539 & 315213\\
 Annotated Frame & 13643  & 16836 & 26440 & 12967 & 11109 & 9109 & 24118 & 114222\\
% Max Events   & 6290 &   5573 & 9022 & 12563 & 7247 & 7646 & 14368 &XXX\\
 Saccade Counts & 406  & 317 & 288 & 317 & 238 & 248 & 287 &2101\\
 Blink Counts & 18 & 3 & 32 & 29 & 14 & 10 & 14 &120\\
 Eye Center Position & 4447 & 2911 & 3400 & 2981 & 2510 & 2917 & 3139 &22305\\
 \hline
 \hline
\end{tabular}
}
\caption{Table reporting core statistics about each annotated user of the Angelopoulos dataset, computed with the proposed method.\label{annotations_statistics}}
\end{table}

In the 114222 annotated frames, it was possible to recognize either an open eye or a blink movement. From these frames, we annotated 22305 distinct eye center positions, detected 2101 saccade movements, and identified 120 blink movements.
All these elements are summarized in Tab~\ref{annotations_statistics}, where it is also possible to see this information data for each annotated user.

These annotations can be used during the training and testing of eye-tracking algorithms that utilize event-camera data to also detect various types of eye movements, such as blinks and saccades.

%% file: content/5-conclusions.tex
\label{sec::end}
With this work, we presented an improved version of the original dataset proposed by Angelopoulos in~\cite{angelopoulos2020event}, together with the pipeline used to generate improved annotations at 200Hz, providing what we believe could be a valuable resource for the scientific community.

The semi-automatic pipeline presented provides a method that reduces both the time required to annotate an event-camera dataset and the potential human error that might arise from a completely manual annotation process. This method can also be applied to future datasets, giving more precise annotations that can help develop and test new algorithms for event-based eye-tracking.

Simultaneously, with the increasing interest in event-based technologies, particularly in low-power applications, we aim to expedite research in real-time eye-tracking using purely event-driven approaches.

\begin{comment}
\[
\mathbf{X} =
\begin{bmatrix}
x \\
y \\
\dot{x} \\
\dot{y}
\end{bmatrix}
\]

\[
\mathbf{F} =
\begin{bmatrix}
1 &  0 & \Delta t & 0 \\
0 & 1 & 0 & \Delta_t\\
0 & 0 & 1 & 0\\
0 & 0 & 0 & 1
\end{bmatrix}
\]

\[
\mathbf{H} =
\begin{bmatrix}
1 & 0 & 0 & 0\\
0 & 1 & 0 & 0\\
\end{bmatrix}
\]

\[
\mathbf{Q} =
\begin{bmatrix}
1 & 0 & 0 & 0\\
0 & 1 & 0 & 0\\
0 & 0 & 1 & 0\\
0 & 0 & 0 & 1
\end{bmatrix}
\]

\[
\mathbf{R} =
\begin{bmatrix}
1 & 0\\
0 & 1\\
\end{bmatrix}
\]

\[
\mathbf{B} = 0
\]
\end{comment}